\documentclass[11pt]{article}
\usepackage{coling2020}
\usepackage{times}
\usepackage{url}
\usepackage{latexsym}
  \usepackage[utf8]{inputenc} 
\usepackage{times}

\usepackage{microtype}

\usepackage{graphicx}

\colingfinalcopy 

\newcommand{\verdd}{Ve\textsuperscript{$\prime$}rdd}

\title{\verdd{}. Narrowing the Gap between Paper Dictionaries,\\Low-Resource NLP and Community Involvement}

\author{Khalid Alnajjar~~~Mika Hämäläinen~~~Jack Rueter~~~Niko Partanen\\
Department of Digital Humanities\\ 
University of Helsinki and Rootroo Ltd \\
\tt firstname.lastname@helsinki.fi}

\date{}

\begin{document}
\maketitle
\begin{abstract}
We present an open-source online dictionary editing system, \verdd{}, that offers a chance to re-evaluate and edit grassroots dictionaries that have been exposed to multiple amateur editors. The idea is to incorporate community activities into a state-of-the-art finite-state language description of a seriously endangered minority language, Skolt Sami. Problems involve getting the community to take part in things above the pencil-and-paper level. At times, it seems that the native speakers and the dictionary oriented are lacking technical understanding to utilize the infrastructures which might make their work more meaningful in the future, i.e. multiple reuse of all of their input. Therefore, our system integrates with the existing tools and infrastructures for Uralic language masking the technical complexities behind a user-friendly UI.
\end{abstract}

\section{Introduction}
\blfootnote{
    %
    %
    %
    %
    %
    %
     \hspace{-0.65cm}  
     This work is licensed under a Creative Commons 
     Attribution 4.0 International License.
     License details:
     \url{http://creativecommons.org/licenses/by/4.0/}.
}



We present an open-source dictionary editing tool\footnote{https://akusanat.com/verdd \\ Source code available: https://github.com/mokha/verdd} called \verdd{}\footnote{\verdd{} means stream in Skolt Sami}. The tool has been and currently is under active development to cater for the needs of Skolt Sami (\textit{ISO 639-2: sms}) speaking language community and their on-going project on modernizing a Finnish-Skolt Sami paper dictionary (see \cite{alnajjar2020editing}). Although Skolt Sami is severely endangered with its 300 native speakers \cite{moseley_2010}, a great deal of NLP tools have been developed for it over the past decade; such as finite-state based morphological analysers and generators in the GiellaLT repository \cite{Moshagen2014}, XML and MediaWiki based online dictionary \cite{rueter2017synchronized} and most recently a universal dependency treebank \cite{b979f78e138246bb8f3b64d40bf2b09f}. However, due to the pluricentric nature of the language (see \cite{SkoltPluric2019}), these tools are far from perfect. One of the core design principles of \verdd{} is to bring these tools closer to non-technical community members editing a high-quality dictionary.

Building dictionaries is an essential part of resource creation when working on endangered low-resource languages. At the same time, lexical resources are an important part of the work done on computational morphological descriptions, such as finite state transducers. We argue that these lines of work have not traditionally entirely met each others. Traditionally the distinction may have been easier, as some dictionaries were intended to be printed, and others served computational infrastructure such as spell checkers. Nowadays, however, all dictionaries are born digital. Much of the dictionary writing work, often connected to traditional linguistic descriptions and the needs of the communities themselves, is still customarily done by hand using ordinary text processing software. 

In other contexts, various other tools have been used. SIL FieldWorks \cite{baines2009a} has been popular among many language documentation projects, although it clearly is not suitable for all projects and lacks many functionalities \cite{rogers2010review}. Commercial tool TLEx \cite{joffe2004tshwanelex} has also been used, although we personally would not prefer attempts at the use of commercial proprietary software in a language documentation context. These also all represent traditional, installed software that do not allow easy cooperation on a larger team level. A project that comes closer to our work is Lexonomy \cite{mvechura2017introducing}. A central difference here is that our work connects the formal computational descriptions to the dictionary editing process, whereas other projects seem to principally offer a digital environment for the traditional dictionary making itself.

Besides editing dictionaries, one important purpose of \verdd{} system is to allow combining information from different dictionaries. Many parts of lexical information that we want to present combines various sources. For example, etymological data by definition involves several dictionaries and their intercomparison. Similarly dialect dictionaries are inherently connected to the lexicons of their corresponding standard languages. 

In many cases such specialized dictionaries may be practical to represent as distinct works, but still their connections to the other resources are myriad, and essential for the whole enterprise. \verdd{} makes it possible to add these relationships between different entry and relation types. Resulting specialized dictionaries can, if wanted, be exported, but this way we avoid repeating the shared parts of the entries and can minimize duplicate efforts.

\section{\verdd{} System}
In this section, we describe the major features implemented in \verdd{}. \verdd{} is developed in Python using Django framework. Django has been picked as it scores high when compared to other web frameworks in terms of quality attributes \cite{plekhanova2009evaluating}. 

When building \verdd{}, modularity was constantly kept in mind to allow the system to be extended, incorporated into other systems or used for other languages. Currently, the system keeps track of the following elements in a dictionary: 1) lexemes, 2) their inflectional paradigms, 3) any relevant external links to them, 4) relations between two lexemes, 5) sources that backup these relations (e.g. other existing dictionaries), and 6) examples and 7) metadata to lexemes and relations. Nonetheless, we are considering adding dialectal transcriptions and locale information to lexemes, which, in addition to preserving this information, would support geolinguistics studies of these languages and facilitate developing computational models for processing dialects (c.f.  \cite{537e4ad5fc054610a23474d91532bb07}).

\verdd{} supports importing existing dictionaries in XML and CSV formats or from the Akusanat MediaWiki dictionary \cite{euraleksi} directly, this is to allow a smooth transition for editors to the tool without the need to input the data manually. In the import process, \verdd{} takes care of wrong character encoding by mapping wrong variations into correct versions. This unification of characters is important as many of the special characters used in Skolt have either emerged in the Unicode standard recently, have wrong, similar looking Unicode characters or are impossible to type without an appropriate keyboard layout. This has lead to a high degree of inconsistencies of the characters used to write Skolt Sami, even if the text has been saved in UTF-8.

\begin{figure}[!htb]
\center{\includegraphics[width=0.48\textwidth]
{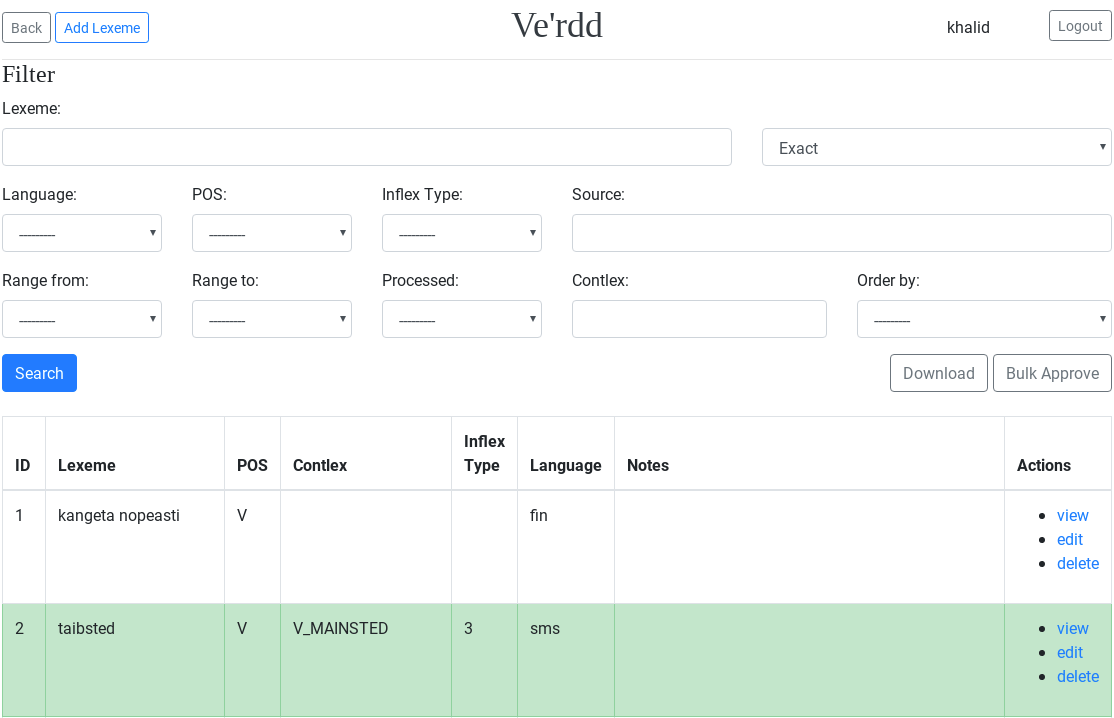}}
\caption{\label{fig:verdd-homepage}The advanced search interface for finding lexemes to be processed.}
\end{figure}

Figure~\ref{fig:verdd-homepage} shows the front page of \verdd{}, in which users can use the advanced search functionality to filter lexemes by lemma (fully, partially or matching a regular expression), language, source they appeared in, whether they have been verified and so on. Additionally, they can sort the result by their assonance and consonance which could help in discovering lexemes sharing an inflectional form. Users can access, edit or delete lexemes from this page. Furthermore, users can download the entire result of the search query or enter the bulk approving mode where they can tick a checkbox to confirm that the information associated with the lexeme is correct, which will highlight the approved lexemes in green as illustrated in the figure. A similar search interface also exists for relations.

\verdd{} utilizes the Skolt FST \cite{rueter2020fst} through UralicNLP \cite{uralicnlp_2019} to produce inflectional paradigms. The transducers are built on HFST \cite{linden2013hfst}, which makes it easy to integrate transducers for other languages as well. The most common paradigms are displayed under the mini-paradigms section; nonetheless, users can access the full list of generated word inflections by clicking on the ``See all miniparadigms'' button. Users have the ability to add new inflectional forms and, in case of a wrong inflection produced by the transducers, they can correct it by adding a form that overwrites the wrong word form. Corrections of this nature are monitored closely and used as a feedback to update the transducers. 

The system organizes the lexicographic data into a list of lexemes that contain all the relevant information to the lexeme itself (such as inflection, language, part-of-speech) and relations in between two lexemes. The relations (such as derivations, compounds and translations) linked to a lexeme are also shown in the lexeme view interface. Sources (e.g. other dictionaries) that support the defined relation, along with example sentences and metadata that are specific to the relation, are presented alongside the relation. The sources functionality makes it possible to compare the different dictionaries that have been imported into the system.

Users can edit, delete or supply \verdd{} with new information regarding any of its elements (e.g. lexemes, relations, sources \ldots etc). \verdd{} keeps track of all versions of instances along with who changed them and when, to mitigate introducing inaccurate information and losing the ability to revert back to the correct instances. 

Once a user has finished checking or processing a lexeme, they can navigate to the next or previous lexeme using the navigation lists at the sides of the lexeme information. The navigation list depends on the search query the user defined during their filtering phase. This gives them the ability to move form a lexeme to another effortlessly without going back to the search results.

At the end of the editing period of the dictionary, approved relations are automatically exported by \verdd{} into a \LaTeX{} file, which are then included in a modular \LaTeX{} dictionary template. The dictionary template is language independent and renders entries produced by \verdd{} using predefined commands as a part of the template, which yields a full print-ready dictionary that is automatically generated. Editors can manually check and polish the entries to ensure that the document satisfies the editorial requirements set for publishing the dictionary.

\section{Catering to the Language Community}

Interaction with members of the language community in charge of editing the dictionary has been an important part of the project since its beginning. In this section, we describe the needs that were identified when discussing with the community members and observing their workflow.

\subsection{Initial Requirements for the System}

As a part of the project of editing a new version of the Finnish-Skolt Sami dictionary, a need for an editing system arose. Since dictionary editing for Skolt Sami has been done either with paper dictionaries in mind or with online dictionaries in mind (c.f. \cite{euraleksi}), a system with a user-interface and functionality supporting both modalities was needed.

Members of any given language community cannot be expected to have mastered language documentation, nor can they be expected to posses the technical skills needed to run command line applications for morphological analysers or edit XML-formatted dictionaries. The system should therefore provide a graphical user interface that can be used simultaneously by multiple non-technical dictionary editors.

An abstraction of the workflow is the following: the dictionary editors go through existing lexicographic resources imported into the system. They need to verify and correct each entry with the possibility of adding new entries when needed. As similar words behave in similar ways, the editors need a mechanism of filtering and sorting the words in the system based on similar vowels (assonance), consonants (consonance) and word ending. For this purpose, \verdd{} has an extensive searching, filtering and sorting functionality.

As editors go through the lexical entries in the system, a history of changes should be kept. \verdd{} includes a special administrator view that shows all the edits done in the system and their respective editors. Edits can be reverted back for individual words or individual relations without the need of reverting anything more than necessary.

Finally, the system should be able to output its data in meaningful formats. This means outputting the final dictionary for printing, a CSV and XML. Some of the dictionary editors are familiar with Excel and they have a need to see the data in a format compatible with the software. Then again, some more technical users are interested in XML for using it for NLP.

The workflow anticipated in the XML, Akusanat and even \verdd{} have, at times, proven to be incompatible with those of the actual native users. This may be the result of experience with pencil and paper approaches to language documentation. Some of the users have been more familiar with ticking translation pairs off in a long list (all on paper first and then on \verdd{}). For this reason we organised sessions with the community members to better understand their needs.

\subsection{First User Session}

The first session with the participating community members was organized in Inari in the Finnish Lapland. Two native Skolt Sami speakers and one non-native Skolt Sami teacher who are to edit the dictionary participated in the tutorial session. The purpose was to get to know better how they do dictionary editing and more concretely what their needs are. This session revealed that several key features were lacking and that the user interface needed more refining for a better usability.

The development language of \verdd{} has been English and therefore the user interface was initially in English. The community members demanded it be localized in Finnish as they are not fluent enough in English to use the system. Another interface problem was that the community members needed a quick visual way of seeing which words and relations they had already verified. Although the system kept track of this already, this was made visually clearer by coloring the words and relations that had already been verified entirely in light green.

By observing how the system was used, we quickly noticed that the editors were consulting several different pages to get their work done. They used Akusanat\footnote{https://www.akusanat.com} to see the full inflectional paradigms of the Skolt Sami words. \verdd{} initially included only a miniparadigm that highlighted only the linguistically meaningful inflections. As the community members are no linguists, however, they felt a need to see the entire full inflection paradigms. This feature was automatically introduced in \verdd{} by inflecting the words with UralicNLP. Simultaneously a feature for editing the paradigms was also introduced in case the FSTs were producing incorrect inflectional forms.

Another website the editors consulted was Sami TermWiki\footnote{https://satni.uit.no/termwiki/}, which contains a list of terms that have been established as the official recommendations by the Sámi Giellagáldu institution. We collected the Skol Sami terms from the Sami TermWiki and added them to \verdd{}. For the words that are recommended by Sámi Giellagáldu, a link to TermWiki appears in \verdd{}.

Two new relation types were requested by the community members. First, they needed to keep some words in the dictionary, although they are not recommended forms, but they need to be kept for the sake of completeness with a reference to the normative form. This relation type was introduced as alternative form relation. Furthermore, there was a wish to link derivational forms to the word they derived from. This was done automatically with the GiellaLT transducers in UralicNLP. We processed all the words in the system and linked the ones that received a derivational morphological reading with a matching lemma and part-of-speech.

\subsection{Second User Session}

The second user session was arranged over Skype with two language community members, a student and the instructor of the dictionary project. In this session, it became evident that the editors had resorted to a more traditional Finnish lexicographic approach, i.e. doing editorial work in a pencil and paper fashion. In keeping with this tradition, the three editors had been directed to inspect lists of Skolt Sami verbs with their Finnish translations by their instructor, head editor.

This workflow, although counter-intuitive from the tool developers' perspective, sits well with the editors. In fact, it is difficult to entice them to use the \verdd{} tool directly; since the previous session only 28 entries with all relations had been approved. Needless to say, another set of printed word lists was requested. The editors preferred a list of words on paper to individual words, one at a time. As one of the editors described it: ``When I pack my suitcase, I don't put in one individual thing at a time, so I don't feel good about dealing with words on an individual basis.'' This, in fact, illustrates the practice where editors want to deal with one set of words at a time, i.e. there might be a part-of-speech constraint or even features of assonance or consonance utilized in the sorting of several words for bulk approval in a dictionary editing system.

We recognized the alignment of linguistic and first language user intuition. While a linguistic approach to inflection type categorization might include bulk assessment of similar assonance or consonance, the native language speakers were also looking for word form associations. For this reason we decided there had to be an easy way to print out a list of source-language and target-language word pairs; a structure which could also be realized as paired words with an adjacent column of tick boxes as well a columns for identification of the individual relation. This latter feature could then be used with feeding the results of pencil-n-paper inspections of translation, derivation and etymology relations. This interface design decision is meant to mimic the experience they would have when using a pencil and a paper, although by using a flat design paradigm as opposed to a fully skeuomorphic design, as there is evidence of the former resulting in a higher perceived usability \cite{spiliotopoulos2018comparative}.

A second feature requested was the ability to add relations more freely to a newly added lemma in addition to simple translation relation. This requires exposing the features stored in the relation information to an editable form in the user interface.

\section{Future Directions and Discussion}

In this paper, we have presented \verdd{}, a dictionary editing system for Skolt Sami. Our system relies on technologies that exist in the exact same format for multiple minority languages in the GiellaLT system. This means that the system can readily be used with a little to no configuration just by adding a new language code from the list of 32 languages currently supported by UralicNLP.

Currently, the system is capable of automatically generating morphological inflections, and these inflectional forms can be edited together with the continuation lexicon information. In other words, this can be used to fix any issues that are present in the FSTs. However, at the moment, this is a manual endeavor. Whenever the inflectional forms are edited, the person in charge of writing the FSTs can see the edits in the administration view of \verdd{} and adjust the FSTs accordingly. A future solution would be to make it possible to inspect and edit FSTs directly in the system, similarly to the system proposed by \cite{lepp-etal-2019-visualizing}.

As a longer term goal for the system is a closer integration with the GiellaLT infrastructure and Akusanat MediaWiki dictionary. \verdd{} currently uses the tools and lexicographic information coming from these systems, but any edits made in \verdd{} do not get reflected back to the other systems. As the focus is currently in finalizing the printed Skolt Sami dictionary, this bi-directionality has been left for the future.

The development of \verdd{} continues in close collaboration with the Skolt Sami language community. The immediate next step is to come into an agreement on the layout of the final paper dictionary. Currently, \verdd{} does support outputting the lexicographic data into a \LaTeX{} template that can be edited before the final PDF version. However, the actual final layout is to be decided.

\bibliographystyle{coling}
\bibliography{acl2020}

\end{document}